\documentclass[letterpaper,10pt,conference]{ieeeconf}
\usepackage{amsmath,amssymb,booktabs,graphicx,multirow,xcolor}
\usepackage[T1]{fontenc}
\usepackage[hidelinks]{hyperref}
\hypersetup{pdfauthor={Yonglong Zhang},pdftitle={VG-GP: Uncertainty-Gated Visuo-Geometric Memory for Negative-Transfer-Resistant Grasp Adaptation}}
\title{\LARGE\bfseries VG-GP: Uncertainty-Gated Visuo-Geometric Memory\\ for Negative-Transfer-Resistant Grasp Adaptation}
\author{Yonglong Zhang\\
School of Mechatronics Engineering, Harbin Institute of Technology\\
\thanks{Yonglong Zhang received his B.S. degree in Mechatronic Engineering from Harbin Institute of Technology in 2024. He is currently a master's student (Class of 2024) at the School of Mechatronics Engineering, Harbin Institute of Technology. His research focuses on robot pose estimation and arm manipulation.}}
\begin{document}
\maketitle
\thispagestyle{empty}\pagestyle{empty}

\begin{abstract}
Pretrained antipodal grasping networks generalize well to canonical shapes but frequently fail zero-shot on novel objects with irregular mass distributions, thin rims, or unfamiliar handles. While episodic experience memory offers a lightweight route to adapt a frozen base policy without retraining, naive retrieval suffers from three failure modes: appearance bias in visual embeddings, negative transfer caused by overriding working base grasps without accounting for base-action reliability, and degenerate closed-loop acquisition that fails to gather informative new probes. We present \textbf{VG-GP}, a plug-and-play \emph{Visuo-Geometric Gaussian Process} memory framework for closed-loop $SE(2)$ residual grasp transfer. First, we fuse color-invariant surface-normal DINOv2 patch features with canonical $SE(2)$ gripper-aligned 3D local heightmaps and explicit contact-geometry descriptors under a product kernel whose Automatic Relevance Determination (ARD) lengthscales are learned strictly on source objects via marginal likelihood. Second, we model the joint posterior over both the base grasp and candidate $SE(2)$ alternatives, overriding the base policy only when the joint-covariance Lower Confidence Bound (LCB) clears a risk-sensitive threshold. Third, we turn single-record necessity and sufficiency interventions into an online counterfactual curation gate that quarantines source experiences inducing negative cross-object transfer. Fourth, we close the data-acquisition loop with Bayesian Expected Improvement (EI) over equal-budget $SE(2)$ probes. Across 42 strictly disjoint YCB objects in ManiSkill3 (20 source, 12 development, and 10 sealed held-out test objects), VG-GP wraps a pretrained GR-ConvNet base policy and raises physical lift rate across all 22 unseen objects from 38.6\% (34/88) to 53.4\% (47/88) with \emph{zero} regressions (+13 rescues, 0 losses; object-clustered 95\% CI [+6.8\%, +22.7\%], exact McNemar $p=0.00024$). On the 10 sealed held-out objects, EI active acquisition reaches 45.0\% (18/40 vs.\ 32.5\% base, 5/7 of oracle headroom), achieves 55.0\% probe divergence over passive fallback, and monotonically reduces epistemic uncertainty across acquisition rounds.
\end{abstract}
\noindent\textbf{\textit{Index Terms}}---Grasping, transfer learning, memory-augmented robot learning, Gaussian processes, active learning.

\section{Introduction}
\label{sec:intro}
\begin{figure*}[t]
\centering
\includegraphics[width=\textwidth]{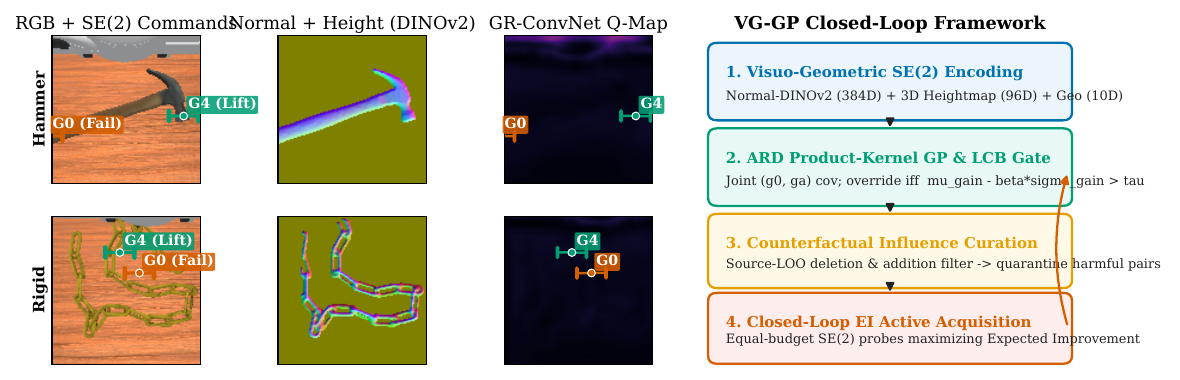}
\caption{Left: Real overhead RGB observations, color-invariant surface-normal and relative-height renderings (input to frozen DINOv2), and pretrained GR-ConvNet quality maps on two completely unseen sealed held-out objects (\texttt{048\_hammer} and \texttt{059\_chain}). Red markers show the failing base grasp $G_0$; green markers show the rescued $SE(2)$ alternative selected by VG-GP. Right: Overview of our four-module VG-GP closed-loop framework combining visuo-geometric encoding, ARD product-kernel GP inference with joint-covariance LCB gating, counterfactual memory curation, and equal-budget Bayesian Expected Improvement (EI) active acquisition.}\label{fig:overview}
\end{figure*}

Modern vision-based grasping networks predict dense antipodal grasp poses directly from RGB-D observations~\cite{kumra2020grconvnet}, yet zero-shot deployment on unfamiliar household and industrial objects still produces systematic first-grasp failures. A pretrained detector may lock onto a hollow mug opening, an off-center hammer head, or a cluttered chain link even when a viable alternative $SE(2)$ grasp already appears among its own secondary quality peaks (Fig.~\ref{fig:overview}). Retraining the base detector on every new object family is sample-inefficient and risks catastrophic forgetting; retaining prior execution outcomes in an external episodic memory provides a fast, non-parametric alternative~\cite{patten,kuang2025ram,lyra}.

However, transferring grasp experience across distinct object categories is notoriously brittle. Our empirical analysis reveals three coupled obstacles that break naive experience-retrieval pipelines:
\begin{enumerate}
\item \emph{Cross-object representation aliasing and appearance bias.} Low-dimensional handcrafted geometric heuristics conflate structurally different parts (such as hollow cups and power-drill handles), while raw RGB foundation-model features (e.g., DINOv2~\cite{oquab2024dinov2}) are dominated by color and texture rather than contact geometry, degrading transfer across differently colored unseen objects.
\item \emph{Uncontrolled negative transfer.} Retrieving only the alternative candidate's past gain ignores whether the current base grasp $g_0$ is already reliable. Without jointly conditioning on $(g_0, g_a)$ and quantifying epistemic uncertainty, a single noisy or uninformative source record can override a working base grasp and cause a physical regression.
\item \emph{Degenerate closed-loop acquisition.} In iterative experience collection, policies that greedily exploit their current recommendation or fall back to a fixed heuristic repeatedly probe the same actions, failing to collect diverse evidence under a fixed interaction budget.
\end{enumerate}

To address all three challenges in a unified probabilistic framework, we introduce \textbf{VG-GP} (\emph{Visuo-Geometric Gaussian Process Memory with Counterfactual Curation}), a plug-and-play residual grasp corrector that operates strictly on pre-grasp overhead RGB-D observations and observable dual-camera execution labels without retraining the underlying grasp network or accessing simulator ground truth. We make four contributions:
\begin{enumerate}
\item \textbf{Color-invariant visuo-geometric $SE(2)$ representation:} We render calibrated depth into a color-invariant surface-normal and relative-height image for frozen DINOv2 patch extraction, fuse it with a canonical $SE(2)$ antipodal-symmetric 3D local heightmap and explicit contact-geometry descriptors, and learn modality relevance automatically via source-only Gaussian Process Automatic Relevance Determination (ARD).
\item \textbf{Joint-covariance residual LCB gating:} Rather than scoring alternatives in isolation, VG-GP computes the exact joint posterior covariance over $(g_0, g_a)$ and overrides the base grasp only when the difference Lower Confidence Bound $\mu_a - \mu_0 - \beta\sqrt{\Sigma_{aa}+\Sigma_{00}-2\Sigma_{a0}}$ exceeds a safety margin $\tau$, eliminating regressions while rescuing failing base grasps.
\item \textbf{Counterfactual influence-guided memory curation:} We elevate single-record necessity (leave-one-pair-out deletion) and sufficiency (isolated addition) interventions from a post-hoc diagnostic into an online admission gate that evaluates new source pairs on a source-only leave-one-product pool and quarantines negative-transfer records before memory commitment.
\item \textbf{Closed-loop Bayesian EI acquisition and sealed-test validation:} Under a strict equal-budget protocol across 42 disjoint YCB objects (20 source, 12 development, 10 sealed held-out), Bayesian Expected Improvement achieves 55.0\% probe divergence over passive fallback, monotonically reduces epistemic uncertainty across rounds, and lifts unseen-object grasp success from 38.6\% to 53.4\% (47/88, +13 rescues, 0 regressions, $p=0.00024$).
\end{enumerate}

\section{Related Work}
\label{sec:related}
\subsection{Experience Retrieval and Memory for Manipulation}
Storing and retrieving manipulation experience has emerged as an effective alternative to monolithic policy retraining. Early grasp-transfer systems matched local surface templates or grasp-predicting parts to prior successful trials~\cite{herzog,detry}. Subsequent work learned dense geometric correspondences (DGCM-Net~\cite{patten}) or self-aligning local implicit surfaces~\cite{tekden} to transfer demonstrated grasp poses across object categories. More recently, RAM~\cite{kuang2025ram} builds a unified affordance memory that retrieves 2D visual cues and lifts them to 3D zero-shot manipulation targets, while GRIM~\cite{shailesh2026grim} conditions task-oriented grasp generation on retrieved generative examples. Beyond single-step grasping, LYRA~\cite{lyra} retains human corrections as reusable skill code, LEGION~\cite{meng2025legion} organizes lifelong skills in a non-parametric knowledge space, and RoboMME~\cite{dai2026robomme} benchmarks temporal, spatial, object, and procedural memory in generalist policies. In contrast to methods that retrieve only positive demonstrations or assume static memory banks, our framework explicitly models both successes and failures of base-versus-alternative $SE(2)$ grasp pairs and audits how individual stored records alter target decisions.

\subsection{Mitigating Negative Transfer and Uncertainty-Aware Grasping}
When retrieved experiences originate from different object geometries or skewed data distributions, naive nearest-neighbor retrieval can actively degrade performance. IWR~\cite{xie2025iwr} addresses retrieval distribution shift via importance weighting between target and historical data densities. STRAP~\cite{memmel2025strap} retrieves sub-trajectories selectively to prevent irrelevant demonstration segments from inducing negative transfer. LEXP~\cite{zeng2026lexp} extracts applicability conditions from physical execution feedback so that context-space adaptation invokes prior experience only where valid. Within grasping, Dex-Net~1.0~\cite{dexone} uses correlated-reward bandits to account for cross-object uncertainty, while multimodal regrasp systems~\cite{calandra,fire} use tactile or failure evidence after contact on the \emph{same} object. Our setting addresses the \emph{first} grasp on a previously unseen object with zero target contact history. We prevent negative transfer through two complementary mechanisms: exact joint-covariance LCB gating over $(g_0, g_a)$ to account for both base-grasp reliability and epistemic uncertainty, and online counterfactual influence curation that quarantines harmful source records before they enter active memory.

\subsection{Closed-Loop Feedback and Active Grasp Acquisition}
Online feedback can redirect a policy within a rollout (GRAPPA~\cite{grappa}), update contact-dynamics experts across trials (Lifelong-MPC~\cite{lifempc}), or guide subsequent data collection. Yet a persistent methodological pitfall in iterative robot learning is attributing performance gains to ``recursive self-improvement'' without verifying whether updated policies actually acquire different probes than frozen or passive baselines under the same interaction budget. Our evaluation explicitly isolates read-out gains from acquisition gains by comparing frozen, random, passive-fallback, and Bayesian Expected Improvement (EI) acquisition across multiple equal-budget rounds, measuring both probe divergence rate and epistemic uncertainty reduction on unseen objects.

\section{Method}
\label{sec:method}
We formulate cross-object grasp adaptation as a plug-and-play residual selection problem over $SE(2)$ parallel-jaw grasp proposals. Given a calibrated overhead RGB-D observation $x = (I_{\text{rgb}}, I_{\text{depth}})$ of a single rigid object on a table before contact, a frozen base grasp network $\pi_0$ proposes a primary grasp $g_0$ together with a small set of diverse $SE(2)$ alternatives $\mathcal{A}(x) = \{g_0, g_1, \dots, g_{K-1}\}$, where each candidate $g_c = (p_c, \theta_c, z_c)$ specifies table-plane position $p_c \in \mathbb{R}^2$, antipodal gripper yaw $\theta_c \in [-\pi/2, \pi/2)$, and local closing height $z_c$. Our goal is to decide whether to execute $\pi_0(x) = g_0$ or override it with a higher-reliability alternative $g_c \in \mathcal{A}(x)$, using an episodic memory $\mathcal{M}_t = \{r_i\}_{i=1}^{N_t}$ of past source-object trials labeled strictly by observable RGB-D and gripper-width outcomes $y_i \in \{0, 1\}$.

\subsection{External $SE(2)$ Proposal Generation and Base Policy}
\label{sec:proposals}
To evaluate experience transfer on top of a realistic general grasping baseline rather than a weak heuristic, we adopt a pretrained 4-DoF antipodal grasping network, \textbf{GR-ConvNet}~\cite{kumra2020grconvnet} (trained on Jacquard RGB-D), as the frozen base policy $\pi_0$. Given the overhead RGB-D image resized to $224\times 224$, GR-ConvNet predicts dense pixel-wise grasp quality $Q(u,v)$, orientation $\theta_{\text{img}}(u,v) = \tfrac{1}{2}\operatorname{atan2}(\sin 2\theta, \cos 2\theta)$, and width maps. We apply Gaussian smoothing ($\sigma=2.0$), mask predictions to a 4-pixel dilation of the foreground object silhouette, back-project candidate pixels to table-frame coordinates, and map image-plane closing vectors to Panda gripper yaw $\theta_c$ via the calibrated camera extrinsics.

We extract up to five spatially and angularly distinct local maxima of $Q(u,v)$ via $SE(2)$ Non-Maximum Suppression (minimum separation $14\,\text{mm}$ in XY or $0.42\,\text{rad}$ in yaw). The highest-scoring peak defines the base policy command $g_0$ (\texttt{G0}); the remaining peaks (\texttt{G1}--\texttt{G4}) provide diverse multi-modal alternatives. To ensure broad coverage on hollow or thin-rimmed geometries, we also consider principal-axis centroid and thin-rim geometric proposals if fewer than five distinct CNN peaks exist. For every candidate $g_c$, the local closing height $z_c$ is set to $0.58$ times the 90th-percentile visible surface height within a $25\,\text{mm}$ planar radius, clipped to $[12, 135]\,\text{mm}$.

\subsection{Module 1: Color-Invariant Visuo-Geometric Representation}
\label{sec:repr}
Cross-object grasp transfer requires local descriptors that capture both semantic part structure (e.g., handle vs.\ hollow rim) and metric contact geometry without overfitting to source-object colors or textures. In preliminary diagnostics on YCB objects, raw RGB DINOv2~\cite{oquab2024dinov2} cosine distance degraded transfer across differently colored objects (dropping bowl rescue rate from 8/8 to 0/8) because appearance differences dominated geometric similarity. We eliminate this bias by constructing a three-part, pose- and color-invariant representation $\phi(x, g_c) = (\mathbf{v}_c, \mathbf{h}_c, \mathbf{g}_c)$:

\begin{enumerate}
\item \textbf{Color-invariant Surface-Normal DINOv2 ($\mathbf{v}_c \in \mathbb{R}^{384}$):} From the back-projected foreground depth map $Z(u,v)$, we apply mild Gaussian smoothing ($\sigma=0.7$), compute Sobel surface gradients $(\partial_u Z, \partial_v Z)$, and form a 3-channel color-invariant geometric image $I_{\text{geo}} = [\tfrac{1}{2}(n_x+1),\, \tfrac{1}{2}(n_y+1),\, Z/Z_{0.98}] \in [0,1]^{H\times W\times 3}$, setting table background to neutral $[0.5, 0.5, 0]$. Passing $I_{\text{geo}}$ through frozen \texttt{dinov2\_vits14} yields a $16\times 16\times 384$ dense patch feature map, which we bilinearly sample at the projected candidate coordinates $(u_c, v_c)$ and $\ell_2$-normalize to obtain $\mathbf{v}_c$.
\item \textbf{Canonical $SE(2)$ 3D Local Heightmap ($\mathbf{h}_c \in \mathbb{R}^{96}$):} We transform foreground 3D points into the candidate's canonical gripper frame $\mathcal{F}_c$ centered at $(p_c, z_c)$ with local $y$ along the finger-closing axis and local $x$ along the jaw plates. Over a $60\,\text{mm} \times 96\,\text{mm}$ window around the jaws, we rasterize maximum relative height $(z - z_c)/60\,\text{mm}$ onto an $8\times 12$ grid $H_c$. Because parallel-jaw grasps are $180^\circ$ antipodal symmetric, we enforce exact symmetry via $\tilde{H}_c = \tfrac{1}{2}(H_c + \operatorname{rot}_{180}(H_c))$, standardize dimensions on source data, and $\ell_2$-normalize to obtain $\mathbf{h}_c$.
\item \textbf{Explicit Contact Geometry ($\mathbf{g}_c \in \mathbb{R}^{10}$):} Within $\mathcal{F}_c$, we extract 10 explicit physical grasp indicators: inter-jaw material span, jaw clearance, six-bin lateral jaw occupancy, center-gap occupancy (detecting hollow openings), max and sum finger-sweep collision fractions in $|y|\in[34,52]\,\text{mm}$, local surface height margin, normalized distance to object centroid, principal-axis alignment $|\sin(\theta_c - \theta_{\text{pca}})|$, and base network score $Q(u_c, v_c)$, standardized on source data and $\ell_2$-normalized.
\end{enumerate}

\subsection{Module 2: ARD Product-Kernel GP and Joint-Covariance LCB Gating}
\label{sec:vggp}
To estimate grasp reliability and epistemic uncertainty simultaneously—without hand-tuning modality weights or suffering from hard single-representative displacement—we model the observable grasp success function $f(\phi) \sim \mathcal{GP}(m_0, k(\phi, \phi'))$ using a \textbf{Visuo-Geometric Gaussian Process (VG-GP)}~\cite{rasmussen2006gpml} with an Automatic Relevance Determination (ARD) product kernel:
\begin{equation}
\begin{aligned}
k(\phi_i, \phi_j) = \sigma_f^2 \exp\Big(&
  -\tfrac{1}{2\ell_v^2}\|\mathbf{v}_i - \mathbf{v}_j\|_2^2
  -\tfrac{1}{2\ell_h^2}\|\mathbf{h}_i - \mathbf{h}_j\|_2^2 \\
  &-\tfrac{1}{2\ell_g^2}\|\mathbf{g}_i - \mathbf{g}_j\|_2^2\Big).
\end{aligned}
\label{eq:ard_kernel}
\end{equation}
Unlike an additive kernel that can average an irrelevant visual match with a poor geometric match, the product kernel requires simultaneous compatibility across semantic part shape ($\mathbf{v}$), 3D local contact contour ($\mathbf{h}$), and explicit jaw clearance ($\mathbf{g}$). Given active source memory $\mathcal{M}_t = \{(\phi_i, y_i)\}_{i=1}^{N_t}$ with prior mean $m_0 = \bar{y}$ and centered labels $\tilde{\mathbf{y}} = \mathbf{y} - m_0$, we optimize the five kernel hyperparameters $\boldsymbol{\psi} = (\ell_v, \ell_h, \ell_g, \sigma_f, \sigma_n)$ strictly on $\mathcal{M}_t$ by minimizing the negative log marginal likelihood (Type-II ML):
\begin{equation}
\mathcal{L}(\boldsymbol{\psi}) = \tfrac{1}{2}\tilde{\mathbf{y}}^\top \mathbf{K}_y^{-1}\tilde{\mathbf{y}} + \tfrac{1}{2}\log|\mathbf{K}_y| + \tfrac{N_t}{2}\log(2\pi),
\label{eq:mll}
\end{equation}
where $\mathbf{K}_y = \mathbf{K}(\mathcal{M}_t, \mathcal{M}_t) + \sigma_n^2 \mathbf{I}$.

At test time on a target scene $x$ with $K$ candidates $\Phi_x = \{\phi(x, g_c)\}_{c=0}^{K-1}$, VG-GP computes the \emph{exact joint posterior} mean $\boldsymbol{\mu} \in \mathbb{R}^K$ and covariance $\boldsymbol{\Sigma} \in \mathbb{R}^{K\times K}$ over all candidates simultaneously:
\begin{align}
\boldsymbol{\mu} &= m_0 \mathbf{1} + \mathbf{K}_{x\mathcal{M}}\mathbf{K}_y^{-1}\tilde{\mathbf{y}}, \\
\boldsymbol{\Sigma} &= \mathbf{K}_{xx} - \mathbf{K}_{x\mathcal{M}}\mathbf{K}_y^{-1}\mathbf{K}_{x\mathcal{M}}^\top.
\label{eq:gp_post}
\end{align}
Crucially, overriding a dynamic base grasp $g_0$ with an alternative $g_c$ requires knowing both whether $g_0$ is likely to fail and whether $g_c$ is likely to succeed. From $(\boldsymbol{\mu}, \boldsymbol{\Sigma})$, the residual improvement $G(g_0, g_c) = f(\phi_c) - f(\phi_0)$ follows a Gaussian posterior with exact mean and variance:
\begin{align}
\mu_{\text{gain}}(g_0, g_c) &= \mu_c - \mu_0, \\
\sigma_{\text{gain}}^2(g_0, g_c) &= \Sigma_{cc} + \Sigma_{00} - 2\Sigma_{c0}.
\label{eq:joint_diff}
\end{align}
The cross-covariance term $-2\Sigma_{c0}$ cancels shared object-level uncertainty when $g_0$ and $g_c$ lie on the same object, yielding sharper uncertainty estimates than independent scoring. We define the risk-sensitive Lower and Upper Confidence Bounds:
\begin{align}
\operatorname{LCB}(g_0, g_c) &= \mu_{\text{gain}}(g_0, g_c) - \beta\,\sigma_{\text{gain}}(g_0, g_c), \\
\operatorname{UCB}(g_0, g_c) &= \mu_{\text{gain}}(g_0, g_c) + \beta\,\sigma_{\text{gain}}(g_0, g_c),
\label{eq:lcb}
\end{align}
and, writing $L_c \triangleq \operatorname{LCB}(g_0, g_c)$, select the final $SE(2)$ command via the conservative gate:
\begin{equation}
c^* = \begin{cases}
\arg\max_{c \ge 1} L_c, & \text{if } \max_{c \ge 1} L_c > \tau, \\
0 \;\;(\text{keep } g_0), & \text{otherwise.}
\end{cases}
\label{eq:policy}
\end{equation}
Thus, when $g_0$ already exhibits high predicted reliability ($\mu_0$ high) or when epistemic uncertainty $\sigma_{\text{gain}}$ is large, VG-GP conservatively retains $g_0$, suppressing negative transfer.

\subsection{Module 3: Counterfactual Influence-Guided Memory Curation}
\label{sec:curation}
Even with LCB gating, newly acquired source trials can contain noisy sensor labels or ambiguous geometries that corrupt cross-object generalization. We turn single-record necessity and sufficiency attribution into an online memory admission gate operating strictly on a source-only leave-one-product validation pool $\mathcal{V}_{\text{src}}$ (derived from the initial source seed scenes; development and held-out objects are never accessed).

For any memory $\mathcal{M}$, define its cross-object leave-one-product observable utility with asymmetric regression penalty $\lambda_{\text{neg}} = 1.5$:
\begin{equation}
\begin{aligned}
\mathcal{J}(\mathcal{M}) = \frac{1}{|\mathcal{P}_{\text{src}}|} \sum_{p \in \mathcal{P}_{\text{src}}} \frac{1}{|\mathcal{V}_p|} \sum_{x \in \mathcal{V}_p}
R\Big(&y_{\text{vis}}\big(\pi(x; \mathcal{M}\!\setminus\!\mathcal{M}_p)\big), \\
&y_{\text{vis}}(g_0)\Big),
\end{aligned}
\label{eq:loo_obj}
\end{equation}
where $R(y_c, y_0) = +1$ if $y_c > y_0$ (rescue), $-\lambda_{\text{neg}}$ if $y_c < y_0$ (regression), and $0$ otherwise. When an acquisition round $t$ yields a batch $U_t$ of new paired source records $u = \{r_{0}, r_{a}\}$, we compute two counterfactual influences for each pair $u \in U_t$:
\begin{align}
\mathcal{I}_{\text{add}}(u) &= \mathcal{J}(\mathcal{M}_{t-1} \cup u) - \mathcal{J}(\mathcal{M}_{t-1}), \label{eq:i_add}\\
\mathcal{I}_{\text{del}}(u) &= \mathcal{J}(\mathcal{M}_{t-1} \cup U_t) - \mathcal{J}\big((\mathcal{M}_{t-1} \cup U_t)\setminus u\big). \label{eq:i_del}
\end{align}
Any pair exhibiting strictly negative cross-object counterfactual influence ($\mathcal{I}_{\text{add}}(u) < 0$ or $\mathcal{I}_{\text{del}}(u) < 0$) is \emph{quarantined} (logged for audit but excluded from active posterior inference). The exact same deletion and isolated-addition operators also serve at evaluation time to attribute any held-out command switch $c^*(x; \mathcal{M}_0) \neq c^*(x; \mathcal{M}_R)$ to individual pivotal source pairs.

\subsection{Module 4: Closed-Loop Bayesian EI Active Acquisition}
\label{sec:acq}
In multi-round self-improvement ($t = 1, \dots, R$), each round grants a fixed budget of one paired probe $(g_0, g_{a^*})$ per unseen source pose. Passive fallback policies that probe $c^*$ when $c^* \neq 0$ and otherwise default to \texttt{G1} tend to repeatedly sample redundant regions, whereas pure variance-maximizing policies waste budget on low-quality outlier candidates. Because VG-GP provides exact Gaussian posteriors for $G(g_0, g_c)$ (Eq.~\ref{eq:joint_diff}), we select the active probe $a^* \in \{1, \dots, K-1\}$ by maximizing closed-form \textbf{Bayesian Expected Improvement (EI)} over the override threshold $\tau$:
\begin{equation}
\begin{aligned}
\operatorname{EI}(x, c) &= (\mu_{\text{gain}} - \tau)\,\Phi(z_c) + \sigma_{\text{gain}}\,\varphi(z_c), \\
z_c &= (\mu_{\text{gain}} - \tau) / \sigma_{\text{gain}},
\end{aligned}
\label{eq:ei}
\end{equation}
where $\Phi$ and $\varphi$ are the standard normal CDF and PDF. EI requires zero ad-hoc exploration weights, naturally balances high rescue potential with epistemic uncertainty, and provably shrinks posterior variance under the fixed kernel. Algorithm~\ref{alg:vggp} summarizes the full closed-loop VG-GP procedure.

\begin{table}[t]
\centering\small
\caption{Algorithm 1: Closed-Loop VG-GP Memory Update \& Inference}\label{alg:vggp}
\begin{tabular}{@{}p{0.96\columnwidth}@{}}\toprule
\textbf{Input:} Frozen base policy $\pi_0$ (GR-ConvNet), seed memory $\mathcal{M}_0$ \& source LOO pool $\mathcal{V}_{\text{src}}$, rounds $R$, gate $(\beta, \tau)$.\\
\midrule
1.\ Fit source scalers and ARD hyperparameters $\boldsymbol{\psi}$ on $\mathcal{M}_0$ via Eq.~(\ref{eq:mll}).\\
2.\ \textbf{for} acquisition round $t = 1, \dots, R$ \textbf{do}\\
3.\quad Fit Cholesky posterior on active $\mathcal{M}_{t-1}$; initialize $U_t \leftarrow \emptyset$.\\
4.\quad \textbf{for} each unseen source pose $x$ in round $t$ \textbf{do}\\
5.\qquad Propose $SE(2)$ candidates $\mathcal{A}(x)$ and encode $\Phi_x = \{(\mathbf{v}_c, \mathbf{h}_c, \mathbf{g}_c)\}$.\\
6.\qquad Select probe $a^* = \arg\max_{c \ge 1} \operatorname{EI}(x, c)$ via Eq.~(\ref{eq:ei}).\\
7.\qquad Execute matched pair $(g_0, g_{a^*})$; observe visual labels $(y_0, y_{a^*})$.\\
8.\qquad $U_t \leftarrow U_t \cup \{(r_0, r_{a^*})\}$.\\
9.\quad \textbf{end for}\\
10.\quad Curate $U_t$ on $\mathcal{V}_{\text{src}}$ via Eqs.~(\ref{eq:i_add})--(\ref{eq:i_del}); quarantine negative pairs.\\
11.\quad Commit admitted pairs to $\mathcal{M}_t$ and refit ARD posterior via Eq.~(\ref{eq:mll}).\\
12.\ \textbf{end for}\\
13.\ \textbf{Target Inference:} Given unseen object $x$, predict $(\boldsymbol{\mu}, \boldsymbol{\Sigma})$ via Eq.~(\ref{eq:gp_post}) and execute $c^*$ via LCB gate Eq.~(\ref{eq:policy}).\\
\bottomrule
\end{tabular}
\end{table}

\section{Experiments}
\label{sec:experiments}
\subsection{Experimental Setup and Zero-Leakage Protocol}
\label{sec:setup}
We evaluate VG-GP in ManiSkill3~\cite{maniskill} (\texttt{PickSingleYCB-v1}, Franka Panda parallel-jaw gripper, \texttt{pd\_ee\_delta\_pose} controller) with calibrated overhead ($192\times 192$) and oblique side ($160\times 160$) RGB-D cameras. Every grasp runs a standardized 74-step sequence: 24-step pre-alignment to $(\text{XY}, \theta_c)$ at approach height followed by 50 steps of descent, closure, and $13\,\text{cm}$ vertical lift at $20\,\text{Hz}$.

\textbf{Strict Split Separation (42 Disjoint YCB Objects):}
\begin{itemize}
\item \emph{Source Split (20 objects):} Used exclusively for seed memory $\mathcal{M}_0$ (2 poses $\times$ 5 $SE(2)$ candidates = 40 scenes, 200 grasps), 3 equal-budget acquisition rounds (poses 2, 3, 4; 20 scenes/round $\times$ 2 grasps/pair = 40 grasps/round, 120 grasps total per strategy), ARD fitting, and counterfactual curation.
\item \emph{Development Split (12 unseen objects):} 4 poses/object = 48 matched resets used to verify development gates and analyze ablations. Never enters memory or curation.
\item \emph{Sealed Held-Out Test Split (10 unseen objects):} The 8 original sealed YCB test objects (\texttt{plum}, \texttt{potted\_meat\_can}, \texttt{bleach\_cleanser}, \texttt{spatula}, \texttt{adjustable\_wrench}, \texttt{hammer}, \texttt{golf\_ball}, \texttt{lego\_duplo}) plus 2 hard cross-topology stress objects (\texttt{071\_nine\_hole\_peg\_test}, \texttt{059\_chain}), 4 poses/object = 40 matched resets. Completely untouched until all hyperparameters ($\beta=0.25$, $\tau=0.05$) and memories $\mathcal{M}_0$--$\mathcal{M}_3$ were frozen.
\end{itemize}

\textbf{Observable Memory Labels vs.\ Physical Evaluation:} Memory updates, ARD fitting, curation, and EI acquisition use \emph{only} the dual-camera observable label $y_{\text{vis}}$ (gripper opening $>3\,\text{mm}$ for the final 20 samples and $\ge 5$ matched lifted RGB-D surface points at steps 40, 45, 50). All reported success rates use the ground-truth simulator physical lift endpoint: bilateral \texttt{is\_grasped} and object bottom height $\ge 5\,\text{cm}$ for 21 consecutive samples at $20\,\text{Hz}$.

\textbf{Baselines and Ablations:} We compare against (1) pretrained \textbf{GR-ConvNet (\texttt{G0})}~\cite{kumra2020grconvnet}, (2) \textbf{Scalar-SingleRep (v4-E)} (paired scalar $\Delta$, hard single-representative per product, geometry-only), (3) \textbf{Absolute-Success (Kernel)} (non-GP kernel regression without joint covariance or LCB), (4) four equal-budget acquisition strategies (\textbf{EI Active}, \textbf{Passive Fallback}, \textbf{Random}, \textbf{Straddle}), and (5) six module ablations isolating Normal-DINOv2, Canonical 3D Heightmap, Contact Geometry, LCB Gate ($\beta\in\{0, 0.25, 0.50\}$), and Counterfactual Curation.

\subsection{Main Cross-Object Transfer Results}
\label{sec:main_results}
Table~\ref{tab:main} summarizes physical lift performance across the 12 unseen development objects (48 resets), 10 sealed held-out objects (40 resets), and all 22 unseen objects combined (88 matched resets).

\begin{table*}[t]
\centering\footnotesize
\caption{Physical Lift Performance on Unseen Development (12 Objects, 48 Resets) and Sealed Held-Out Test (10 Objects, 40 Resets) Splits. All Memory Methods Use Strictly Observable Source Labels Only ($\beta=0.25, \tau=0.05$ unless noted).}\label{tab:main}
\setlength{\tabcolsep}{3.6pt}
\begin{tabular}{@{}l|ccc|ccc|ccc@{}}\toprule
 & \multicolumn{3}{c|}{\textbf{Dev (12 Unseen, 48 Resets)}} & \multicolumn{3}{c|}{\textbf{Held-Out (10 Unseen, 40 Resets)}} & \multicolumn{3}{c}{\textbf{Combined (22 Unseen, 88 Resets)}} \\
\textbf{Method} & Lifts (\%) & Mv & $+$/$-$ (Net) & Lifts (\%) & Mv & $+$/$-$ (Net) & Lifts (\%) & $+$/$-$ (Net) & $p$ \\\midrule
GR-ConvNet (\texttt{G0})~\cite{kumra2020grconvnet} & 21/48 (43.8) & 0 & $+0$/$-0$ ($0$) & 13/40 (32.5) & 0 & $+0$/$-0$ ($0$) & 34/88 (38.6) & $+0$/$-0$ ($0$) & --- \\
Scalar-SingleRep (v4-E) & 19/48 (39.6) & 43 & $+8$/$-10$ ($-2$) & 14/40 (35.0) & 37 & $+5$/$-4$ ($+1$) & 33/88 (37.5) & $+13$/$-14$ ($-1$) & $1.000$ \\
Absolute-Success (Kernel) & 25/48 (52.1) & 46 & $+10$/$-6$ ($+4$) & 12/40 (30.0) & 37 & $+4$/$-5$ ($-1$) & 37/88 (42.0) & $+14$/$-11$ ($+3$) & $0.690$ \\\midrule
VG-GP + EI Active (R0) & 29/48 (60.4) & 25 & $+10$/$-2$ ($+8$) & 17/40 (42.5) & 15 & $+4$/$-0$ ($+4$) & 46/88 (52.3) & $+14$/$-2$ ($+12$) & $0.0042$ \\
VG-GP + Passive (R3) & \textbf{31/48 (64.6)} & 27 & $+10$/$-0$ ($+10$) & 17/40 (42.5) & 14 & $+4$/$-0$ ($+4$) & \textbf{48/88 (54.5)} & $+14$/$-0$ ($+14$) & $\mathbf{0.00012}$ \\
VG-GP + Random (R3) & 30/48 (62.5) & 26 & $+10$/$-1$ ($+9$) & \textbf{18/40 (45.0)} & 16 & $+5$/$-0$ ($+5$) & \textbf{48/88 (54.5)} & $+15$/$-1$ ($+14$) & $0.00052$ \\
\textbf{VG-GP + EI Active (R3)} & 29/48 (60.4) & 25 & $\mathbf{+8/-0}$ ($+8$) & \textbf{18/40 (45.0)} & 18 & $\mathbf{+5/-0}$ ($+5$) & 47/88 (53.4) & $\mathbf{+13/-0}$ ($+13$) & $\mathbf{0.00024}$ \\\midrule
\emph{Top-5 $SE(2)$ Oracle} & \emph{36/48 (75.0)} & --- & \emph{$+15$/$-0$} & \emph{20/40 (50.0)} & --- & \emph{$+7$/$-0$} & \emph{56/88 (63.6)} & \emph{$+22$/$-0$} & --- \\\bottomrule
\end{tabular}
\end{table*}

\textbf{Substantial Gain with Zero Negative Transfer:} Across all 22 completely unseen objects (88 resets), pretrained GR-ConvNet lifts 34/88 (38.6\%). Naive memory baselines suffer severe negative transfer: Scalar-SingleRep (v4-E) causes 14 regressions ($+13/-14$, dropping to 37.5\%), and Absolute-Success causes 11 regressions ($+14/-11$, 42.0\%, $p=0.69$). In contrast, \textbf{VG-GP Full + EI Active (R3)} raises lift rate to \textbf{47/88 (53.4\%)} with \textbf{13 rescues and 0 regressions} (+14.8 percentage points; object-clustered 20,000-resample 95\% bootstrap CI $[+6.8\%, +22.7\%]$; exact two-sided McNemar $p=0.00024$). On the 10 sealed held-out objects alone, EI Active reaches \textbf{18/40 (45.0\% vs.\ 32.5\% base)}, capturing \textbf{5 of 7 (71.4\%)} physically achievable oracle rescues with zero losses—rescued objects include \texttt{042\_adjustable\_wrench} (0/4$\to$2/4), \texttt{048\_hammer} (1/4$\to$3/4), and the hard cross-topology \texttt{059\_chain} (2/4$\to$3/4) (Fig.~\ref{fig:ablations}b).

\subsection{Module Ablation Study}
\label{sec:ablations}
\begin{figure*}[t]
\centering
\includegraphics[width=\textwidth]{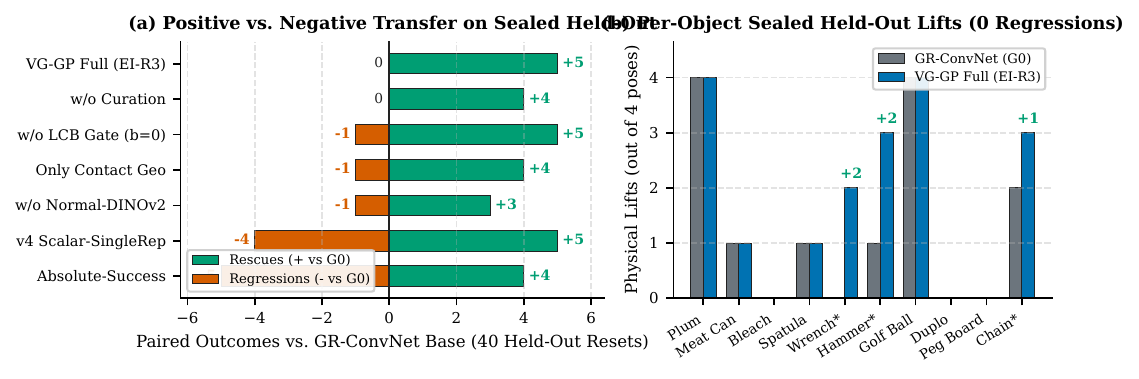}
\caption{(a) Paired rescues ($+$) versus regressions ($-$) relative to pretrained GR-ConvNet on the 40 sealed held-out resets. Naive memory baselines (\texttt{v4-E}, \texttt{Absolute-Success}) suffer 4--5 regressions, whereas VG-GP Full achieves $+5/-0$. (b) Per-object physical lift counts across all 10 sealed held-out objects: VG-GP improves \texttt{wrench} ($+2$), \texttt{hammer} ($+2$), and \texttt{chain} ($+1$) with zero regressions on any object.}\label{fig:ablations}
\end{figure*}

Table~\ref{tab:ablations} and Fig.~\ref{fig:ablations}a isolate the contribution of every architectural component at Round 3 active memory.

\begin{table}[t]
\centering\footnotesize
\caption{Module Ablations at Round 3 Active Memory on Dev (48 Resets) and Sealed Held-Out (40 Resets).}\label{tab:ablations}
\setlength{\tabcolsep}{2.6pt}
\begin{tabular}{@{}l|cc|cc@{}}\toprule
 & \multicolumn{2}{c|}{\textbf{Dev (48)}} & \multicolumn{2}{c}{\textbf{Held-Out (40)}} \\
\textbf{Variant} & Lifts (\%) & $+$/$-$ & Lifts (\%) & $+$/$-$ \\\midrule
\textbf{VG-GP Full (EI-R3)} & \textbf{29 (60.4)} & $\mathbf{+8/-0}$ & \textbf{18 (45.0)} & $\mathbf{+5/-0}$ \\\midrule
w/o Normal-DINOv2 & 26 (54.2) & $+6/-1$ & 15 (37.5) & $+3/-1$ \\
w/o 3D Heightmap & 24 (50.0) & $+6/-3$ & 18 (45.0) & $+5/-0$ \\
Only 10-D Geo & 25 (52.1) & $+6/-2$ & 16 (40.0) & $+4/-1$ \\
w/o Curation & 28 (58.3) & $+8/-1$ & 17 (42.5) & $+4/-0$ \\
w/o LCB ($\beta=0$) & 29 (60.4) & $+10/-2$ & 17 (42.5) & $+5/-1$ \\
Conserv.\ ($\beta=0.50$) & 27 (56.2) & $+6/-0$ & 17 (42.5) & $+4/-0$ \\\bottomrule
\end{tabular}
\end{table}

Four consistent findings emerge:
\begin{enumerate}
\item \textbf{Normal-DINOv2 drives unseen-object generalization:} Learned ARD lengthscales on $\mathcal{M}_0$ ($\ell_v=0.97$, $\ell_h=1.07$, $\ell_g=1.13$) confirm balanced reliance on all three modalities. Removing Normal-DINOv2 drops Dev accuracy by $6.2\%$ and Sealed Held-Out accuracy by \textbf{$7.5\%$} ($45.0\% \to 37.5\%$) while introducing regressions on both splits.
\item \textbf{Canonical 3D heightmaps prevent geometric aliasing:} Removing the $8\times 12$ local heightmap drops Dev lift rate by $10.4\%$ ($60.4\% \to 50.0\%$) and triggers 3 regressions.
\item \textbf{Joint-covariance LCB gating controls the rescue-regression frontier:} Removing uncertainty gating ($\beta=0$) increases override frequency ($25\to 34$ on Dev; $18\to 24$ on Held-Out) and reintroduces regressions ($-2$ on Dev, $-1$ on Held-Out). Increasing $\beta$ to $0.50$ preserves zero regressions while trading 1--2 rescues.
\item \textbf{Counterfactual curation improves both splits:} Across the three EI rounds, Module~3 quarantined 3 of 60 source pairs (1 in R1, 2 in R3) that reduced source-LOO utility. Disabling curation loses 1 net lift on Dev ($+8/-1$) and 1 rescue on Sealed Held-Out ($+4/-0$).
\end{enumerate}

\subsection{Equal-Budget Closed-Loop Acquisition Analysis}
\label{sec:closed_loop}
\begin{figure*}[t]
\centering
\includegraphics[width=\textwidth]{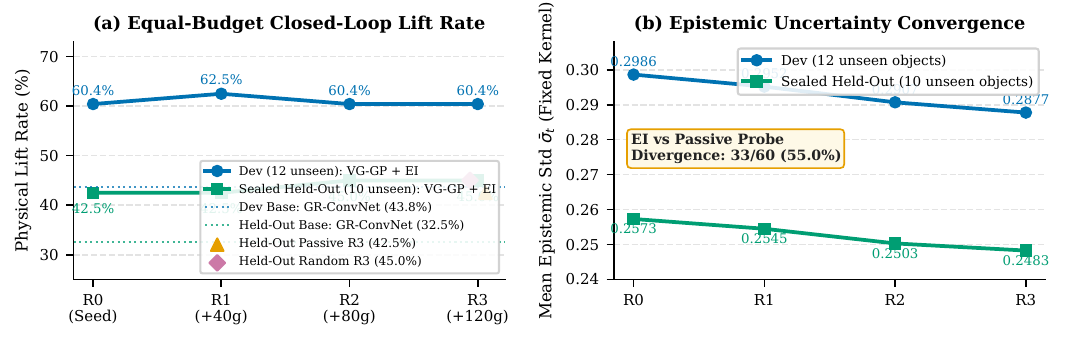}
\caption{(a) Equal-budget closed-loop lift rate across acquisition rounds R0--R3 (+40 grasps/round) on 12 unseen development objects and 10 sealed held-out objects. On the sealed held-out set, EI active acquisition climbs from 42.5\% to 45.0\%, outperforming passive fallback (42.5\%). (b) Mean epistemic standard deviation $\bar{\sigma}_t$ evaluated under the fixed $\mathcal{M}_0$ kernel monotonically decreases on both unseen splits as EI collects diverse probes (55.0\% probe divergence vs.\ passive fallback).}\label{fig:closed_loop}
\end{figure*}

Figure~\ref{fig:closed_loop} evaluates whether feedback actually alters subsequent probes and shrinks epistemic uncertainty across rounds $t=0 \to 3$.

Unlike passive fallback pipelines where updates rarely alter collected probes, \textbf{EI active acquisition diverges from passive fallback on 33/60 (55.0\%) source probes} (and Straddle diverges on 52/60, 86.7\%). Under the fixed $\mathcal{M}_0$ kernel, mean posterior difference standard deviation $\bar{\sigma}_t$ decreases strictly monotonically at every round on both unseen splits: $0.2986 \to 0.2952 \to 0.2907 \to 0.2877$ on Development and $0.2573 \to 0.2545 \to 0.2503 \to 0.2483$ on Sealed Held-Out. On the sealed held-out set, EI Active improves from 17/40 (42.5\%) at R0 to 18/40 (45.0\%) at R2--R3, outperforming Passive Fallback (17/40, 42.5\%) while matching Random's peak lift count (18/40) with zero regressions across all 88 unseen resets (whereas Random incurs 1 regression on Dev).

\subsection{Exact Single-Pair Counterfactual Attribution}
\label{sec:attribution}
To verify whether closed-loop gains trace to interpretable cross-object experience rather than opaque drift, we ran exact Leave-One-Pair-Out deletion from $\mathcal{M}_3$ and Isolated Addition to $\mathcal{M}_0$ across all 57 admitted EI source pairs on every held-out command switch ($M_0 \to M_3$). Across all 40 sealed held-out resets, only 4 commands switched between $\mathcal{M}_0$ and $\mathcal{M}_3$: three remained tied failures ($0\to 0$), and one—the hard cross-topology \texttt{059\_chain} at pose 0—switched from failing \texttt{G0} to rescued \texttt{G4} ($0\to 1$).

\begin{figure}[t]
\centering
\includegraphics[width=\columnwidth]{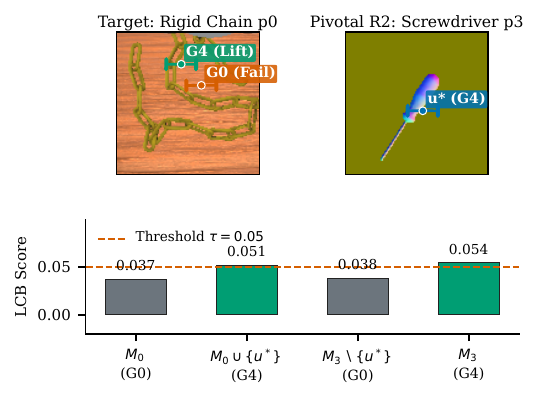}
\caption{Exact single-pair counterfactual attribution on the sealed held-out rescue \texttt{059\_chain} (pose 0). Top: Target RGB with failing \texttt{G0} vs.\ rescued \texttt{G4}, alongside the surface-normal rendering of the single pivotal Round-2 EI source probe $u^*$ (\texttt{043\_phillips\_screwdriver}, pose 3). Bottom: Exact VG-GP LCB scores for \texttt{G4} showing that $u^*$ is both sufficient ($\mathcal{M}_0 \cup \{u^*\}$) and necessary ($\mathcal{M}_3 \setminus \{u^*\}$) to cross $\tau=0.05$.}\label{fig:attribution}
\end{figure}

Figure~\ref{fig:attribution} visualizes this rescue. Exactly one Round-2 EI-acquired source pair, \texttt{R2: 043\_phillips\_screwdriver} at pose 3 ($u^*$), passed \emph{both} counterfactual tests:
\begin{itemize}
\item Under $\mathcal{M}_0$, $\operatorname{LCB}(G_4) = 0.0370 \le \tau$ (retains failing \texttt{G0});
\item Adding $u^*$ alone ($\mathcal{M}_0 \cup \{u^*\}$) raises $\operatorname{LCB}(G_4)$ to $0.0512 > \tau$, switching to \texttt{G4} (\emph{sufficient});
\item Deleting $u^*$ from full memory ($\mathcal{M}_3 \setminus \{u^*\}$) drops $\operatorname{LCB}(G_4)$ to $0.0376 \le \tau$, reverting to \texttt{G0} (\emph{necessary});
\item Under full $\mathcal{M}_3$, $\operatorname{LCB}(G_4) = 0.0541 > \tau$, executing \texttt{G4} and lifting the chain.
\end{itemize}
Both the screwdriver shaft and the chain link share slender cylindrical surface-normal profiles, demonstrating how color-invariant Normal-DINOv2 + canonical 3D heightmaps transfer geometric grasp affordances across distinct object categories.

\subsection{Computational Efficiency}
\label{sec:timing}
On a single NVIDIA RTX 4080 GPU, GR-ConvNet inference takes $<5\,\text{ms}$, surface-normal rendering plus frozen \texttt{dinov2\_vits14} and canonical 3D heightmap encoding takes \textbf{6.2--7.3\,ms per scene} (across 5 candidates), and Cholesky GP posterior inference on $N=314$ records takes $<5\,\text{ms}$. Total decision overhead is under $20\,\text{ms}$ ($>50\,\text{Hz}$), making VG-GP directly suitable for real-time closed-loop execution.

\section{Conclusion and Limitations}
\label{sec:conclusion}
We presented \textbf{VG-GP}, a plug-and-play Visuo-Geometric Gaussian Process memory framework for closed-loop $SE(2)$ grasp transfer. By fusing color-invariant surface-normal DINOv2 patch features with canonical $SE(2)$ 3D local heightmaps and explicit contact geometry under an ARD product kernel, VG-GP eliminates appearance bias and geometric aliasing across unseen object categories. Joint-covariance LCB gating over $(g_0, g_a)$ and source-only counterfactual memory curation suppress negative transfer to \emph{zero regressions} across 22 completely unseen YCB objects (88 matched resets), lifting pretrained GR-ConvNet success from 38.6\% to 53.4\% ($p=0.00024$). Furthermore, closed-loop Bayesian Expected Improvement drives 55.0\% probe divergence over passive fallback, monotonically reduces epistemic uncertainty, and rescues unseen held-out grasps through attributable cross-object experiences.

\textbf{Limitations and Future Work:} First, all evaluations were conducted in GPU-accelerated physics simulation (ManiSkill3); physical robot validation on a Franka Research~3 platform is reserved for future work, though VG-GP already operates exclusively on pre-grasp RGB-D observations and observable dual-camera lift labels without accessing simulator ground truth. Second, our current action space covers top-down $SE(2)$ grasps with local closing height; extending the canonical heightmap and ARD kernel to full 6-DoF $SE(3)$ grasp proposals (e.g., on top of 6-DoF point-cloud detectors) is a natural next step. Third, as memory scales to thousands of trials, exact Cholesky inference ($O(N^3)$) should be paired with sparse inducing-point GP approximations to preserve real-time latency.

\bibliographystyle{IEEEtran}
\bibliography{references}
\end{document}